\documentclass{article}
\usepackage{spconf,amsmath,graphicx,hyperref}
\usepackage{amssymb}
\usepackage{algorithm}
\usepackage{makecell, multirow}
\usepackage{graphicx}
\usepackage{subcaption}
\usepackage{algorithmicx}
\usepackage{algpseudocode}
\usepackage{amsmath}
\usepackage{bm}

\title{Incomplete Depression Feature Selection with Missing EEG Channels}
\name{Zhijian Gong$^{1,*}$\qquad Wenjia Dong$^{1,*}$\qquad Xueyuan Xu$^{1,\dag}$\qquad Fulin Wei$^{2}$\qquad Chunyu Liu$^{3}$\qquad Li Zhuo$^{1}$ \thanks{\hspace{-10pt}*Co-first author, contributed equally to this work  \\ \hspace{-10pt} $^{\dag}$Corresponding author: Xueyuan Xu (xxy@bjut.edu.cn) }}
\address{$^{1}$Beijing University of Technology, Beijing, China \\ $^{2}$Anhui University, Hefei, China \\ $^{3}$North China Electric Power University, Beijing, China}
\begin{document}
\ninept
\maketitle
\begin{abstract}
As a critical mental health disorder, depression has severe effects on both human physical and mental well-being. Recent developments in EEG-based depression analysis have shown promise in improving depression detection accuracies. However, EEG features often contain redundant, irrelevant, and noisy information. Additionally, real-world EEG data acquisition frequently faces challenges, such as data loss from electrode detachment and heavy noise interference. To tackle the challenges, we propose a novel feature selection approach for robust depression analysis, called Incomplete Depression Feature Selection with Missing EEG Channels (IDFS-MEC). IDFS-MEC integrates missing-channel indicator information and adaptive channel weighting learning into orthogonal regression to lessen the effects of incomplete channels on model construction, and then utilizes global redundancy minimization learning to reduce redundant information among selected feature subsets. Extensive experiments conducted on MODMA and PRED-d003 datasets reveal that the EEG feature subsets chosen by IDFS-MEC have superior performance than 10 popular feature selection methods among 3-, 64-, and 128-channel settings.
\end{abstract}


\section{Introduction}

Electroencephalography (EEG) provides rich temporal information through multiple spatial channels, making it a valuable measure for depression analysis \cite{shatte2019machine}. However, the large number of EEG channels often introduces redundant and noise features, which hinders the accuracy of EEG-based depression diagnosis. Feature selection is an effective strategy to address the issues by reducing dimensionality and enhancing interpretability.

\begin{figure}[!t]
    \centering
  %  \hspace{-0.02\textwidth}
    \includegraphics[width=0.498\textwidth]{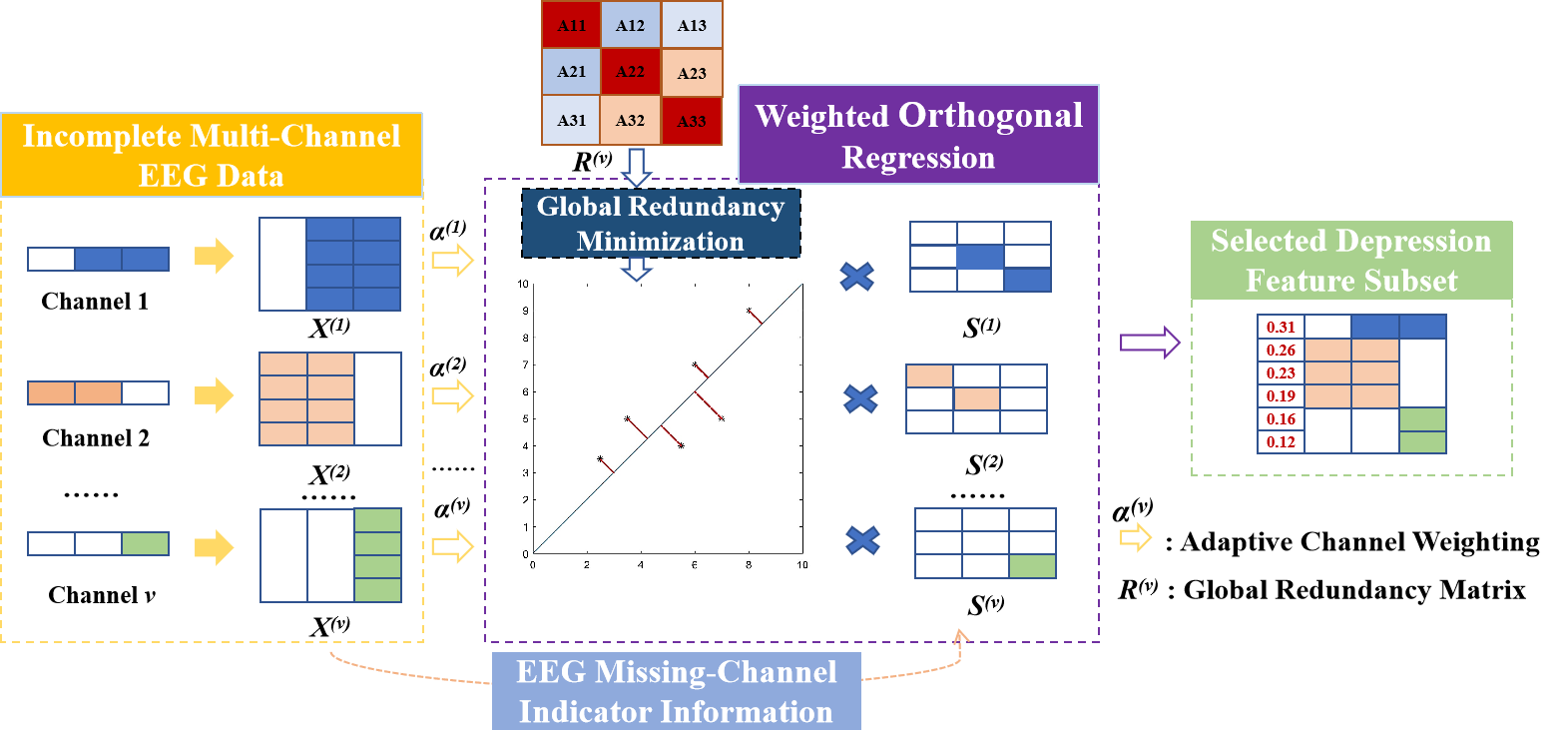}
  %  \captionsetup{font={small}}
    \caption{Overall framework of IDFS-MEC. IDFS-MEC comprises three modules: Weighted orthogonal regression, adaptive channel weighting learning, and global redundancy minimization learning. Missing-channel indicator information is introduced as the weighted matrix for orthogonal regression. IDFS-MEC integrates weighted orthogonal regression and adaptive channel weighting learning to lessen the effects of incomplete channels on model construction, and then utilizes global redundancy minimization learning to reduce redundant information among selected EEG feature subsets.}
  %  \vspace{-5pt}
    \label{Mindmap}
\end{figure}

Feature selection methods are generally categorized into filter, wrapper, and embedded approaches. Filter-based methods apply variable ranking techniques and carry a lower risk of overfitting \cite{theng2024feature}. However, filter methods may neglect inter-variable correlations, which often limits the effectiveness of the selected subsets \cite{bolon2015recent}. Wrapper-based methods evaluate feature subsets based on classifier performance, offering higher accuracy at high computational cost \cite{kuzudisli2023review}. Yet, they are often criticized for being computationally intensive and susceptible to overfitting \cite{kuzudisli2023review}. Compared to filter and wrapper methods, embedded methods integrate feature selection within the model training process, improving efficiency while reducing redundancy \cite{guo2019multi}. Among embedded methods, orthogonal regression–based techniques could retain more discriminative and structural information in high-dimensional data \cite{FSOR2}.

Despite these advantages, the effectiveness of existing approaches is limited by incomplete EEG channel data which often caused by electrode detachment or transmission errors \cite{little2019statistical}. Channel loss occurs frequently in clinical depression detection due to the practical challenges of EEG acquisition \cite{eegIncomplete}. As a result, channel-level data loss can weaken feature representativeness and reduce model accuracy especially in multi-channel EEG data, which ultimately compromising the reliability of depression diagnosis \cite{van2018flexible}.

To address the problems, we propose a novel orthogonal regression based feature selection approach for robust depression analysis, called Incomplete Depression Feature Selection with Missing EEG Channels (IDFS-MEC). The main contributions are as follows:
% \vspace{-3pt}
\begin{itemize}
\item {}\textbf{A novel method for depression identification with incomplete EEG channels:} IDFS-MEC incorporates missing-channel indicator information and adaptive channel weight learning into orthogonal regression to handle incomplete channels. In addition, global redundancy minimization learning is applied to minimize redundancy among feature subsets.
% \vspace{-3pt}
\item {}\textbf{An efficient optimization strategy:} An optimization algorithm integrates Generalized Power Iteration (GPI) \cite{GPI} with general augmented Lagrangian multiplier(ALM) \cite{ALM} to optimize objective function of IDFS-MEC.
% \vspace{-3pt}
\item {}\textbf{Vast empirical validation:}  Extensive experiments were preformed on the MODMA and PRED-d003 datasets across \textbf{3-, 64-, and 128-channels} and the experimental results show that IDFS-MEC outperforms among ten advanced feature selection methods.
\end{itemize}

\section{THE PROPOSED FRAMEWORK}

As shown in Fig. \ref{Mindmap}, raw EEG data with varying missing channels were processed to construct the EEG feature set. Adaptive channel weighting and weighted orthogonal regression were then applied, where missing-channel indicator information incorporated to reduce the impact of channel loss variability on feature selection. Global redundancy minimization learning is utilized to reduce redundant information among selected EEG feature subsets.

\subsection{Global Redundancy Matrix}
A global redundancy matrix $R$ further minimized redundancy of features which can be computed with following expression:  
\begin{equation} 
R_{i,j} = \left( \frac{f_i^T f_j}{\|f_i\| \|f_j\|} \right)^2  
\end{equation} 
where \( f_i \in \mathbb{R}^{n \times 1} \) and \( f_j \in \mathbb{R}^{n \times 1} \) represent column vectors corresponding to the \( i \)-th and \( j \)-th features (\( x_i \) and \( x_j \)), respectively. \( f_i \in \mathbb{R}^{n \times 1} \) and \( f_j \in \mathbb{R}^{n \times 1} \) are calculated using a centering matrix \( Z \) as follows:  

\begin{equation}
\begin{cases}  
f_i = Z x_i^T \\
f_j = Z x_j^T  
\end{cases}  
\end{equation}
in which \( I_n \) is the identity matrix and $Z = I_n - \frac{1}{n}  1_d 1_d{}^T$. 

\subsection{Objective Function}
To achieve effective feature selection with missing channels, the following objective function is constructed:
\begin{equation} 
\begin{aligned}
&\min_{W^{\left( v \right)}, \Theta^{\left( v \right)},  \alpha^{\left( v \right)}, b^{\left( v \right)}} \sum_{v=1}^{\text{ch}}
\left(\alpha^{\left( v \right)}\right)^\gamma 
\left[\lambda \theta^{\left( v \right)}{}^\textsuperscript{\textit{T}} R^{\left( v \right)} \theta^{\left( v \right)} + \right.\\ 
& \left. \left\| \left( W^{\left( v \right)}{}^\textsuperscript{\textit{T}} \Theta^{\left( v \right)} X^{\left( v \right)} + b^{\left( v \right)} 1_{d^{\left( v \right)}}^T - Y \right)S^{\left( v \right)}\right\|_F^2 \right]
\end{aligned} 
\nonumber
\end{equation}
\begin{equation} 
\begin{aligned}
\text{s.t. } & W^{\left( v \right)}{}^\textsuperscript{\textit{T}} W^{\left( v \right)} = I_c, \quad \theta^{\left( v \right)}{}^\textsuperscript{\textit{T}} 1_{d^{\left( v \right)}} = 1, \quad \theta^{\left( v \right)} \geq 0, \\
& \quad \sum_{v=1}^{\text{ch}} \alpha^{\left( v \right)} = 1, \quad 0 \leq \alpha^{\left( v \right)} \leq 1  
\end{aligned} 
\label{Pf1}
\end{equation}  
where $ch$ indicates the channel number and $v$ represents corresponding channel index. $n$, $c$, and $d^{\left( v \right)}$ are the number of samples, classes, and features per channel. $\alpha^{\left( v \right)} \in \mathbb{R}^{d^{\left( v \right)}}$ is a channel weight vector regulated by $\gamma$. \( \lambda \) is a regularization parameter. $\theta^{\left( v \right)}\in \mathbb{R}^{d^{\left( v \right)}}$ is a vector formed by the feature weights and matrix \( \Theta^{\left( v \right)} \) is the diagonalizing result of $\theta^{\left( v \right)}$. 

$R^{\left( v \right)} \in \mathbb{R}^{d^{\left( v \right)} \times d^{\left( v \right)}}$ is the feature redundancy matrix of $v$-th channel. $W^{\left( v \right)} \in \mathbb{R}^{d^{\left( v \right)} \times c}$ is an orthogonal projection matrix.  $b^{\left( v \right)}  \in \mathbb{R}^{c \times 1}$ is a bias vector. $1_{d^{\left( v \right)}}$ represents all-ones vector with $d^{\left( v \right)}$ dimensions. $Y \in \mathbb{R}^{c \times n}$ is the depression label matrix. $X^{\left( v \right)}\in \mathbb{R}^{d^{\left( v \right)} \times n}$ is the incomplete EEG feature matrix with $v$-th channel. $S^{\left( v \right)} \in \mathbb{R}^{n \times n}$ is a missing-channel indicator matrix that the diagonal element $S^{\left( v \right)}_{ii}$ = 0, if the entry $X_{.i}^{\left( v \right)}$ of matrix $X^{\left( v \right)}$ is missing, otherwise $S^{\left( v \right)}_{ii}$= 1.

\section{OPTIMIZATION STRATEGY}
% First, when we take the partial derivative of Eq. (\ref{Pf1}) with respect to $b$ and set it to zero, we have:
% \begin{equation}
% b^{\left( v \right)} = \frac{Y S^{\left( v \right)} S^{\left( v \right)}{}^\textsuperscript{\textit{T}} 1_n - W^{\left( v \right)}{}^\textsuperscript{\textit{T}} \Theta^{\left( v \right)} X^{\left( v \right)} S^{\left( v \right)} }{\text{Tr}(S^{\left( v \right)} S^{\left( v \right)}{}^\textsuperscript{\textit{T}})} 
% \end{equation} 
% By substituting $b$ into Eq. (\ref{Pf1}), we obtain:
By taking the partial derivative with respect to the bias term $b^{\left( v \right)}$, setting it to zero, and substituting it back, the objective function of Eq.~\eqref{Pf1} could be reformulated as:
\begin{equation}
\begin{aligned}
&\min_{W^{\left( v \right)}, \theta^{\left( v \right)}, \alpha^{\left( v \right)}} \sum_{v=1}^{ch} \left(\alpha^{\left( v \right)}\right)^\gamma \left[ \lambda \theta^{\left( v \right)}{}^\textsuperscript{\textit{T}} R^{\left( v \right)} \theta^{\left( v \right)}+ \right. \\
&\left. \| W^{\left( v \right)}{}^\textsuperscript{\textit{T}} \Theta^{\left( v \right)} X^{\left( v \right)} S^{\left( v \right)} H^{\left( v \right)} - Y S^{\left( v \right)} H^{\left( v \right)} \|_F^2\right] 
\nonumber
\end{aligned}
\end{equation}
 
\begin{equation} 
\begin{aligned}
\text{s.t. } & W^{\left( v \right)}{}^\textsuperscript{\textit{T}} W^{\left( v \right)} = I_c, \quad \theta^{\left( v \right)}{}^\textsuperscript{\textit{T}} 1_{d^{\left( v \right)}} = 1, \quad \theta^{\left( v \right)} \geq 0, \\
& \quad \sum_{v=1}^{\text{ch}} \alpha^{\left( v \right)} = 1, \quad 0 \leq \alpha^{\left( v \right)} \leq 1  
\label{fixedb}
\end{aligned}
\end{equation}
%H in Eq. \ref{fixedb} is derived from Eq. \ref{Horigin} , shown bellow: 
where $H^{\left( v \right)} = I_n - \frac{S^{\left( v \right)} \cdot 1_n \cdot 1_n^T}{\text{Tr}(S^{\left( v \right)} \cdot S^{\left( v \right)}{}^\textsuperscript{\textit{T}})}$.

\subsection{Update $W^{\left( v \right)}$ by fixing  $\Theta^{\left( v \right)}$, $\alpha^{\left( v \right)}$}
When $\Theta^{\left( v \right)}$ and $\alpha^{\left( v \right)}$ are fixed and irrelevant terms are omitted, the following function for $W^{\left( v \right)}$ can be written as:

\begin{equation}
\min_{W^{\left( v \right)}{}^\textsuperscript{\textit{T}} W^{\left( v \right)} = I_c} \mathrm{Tr}(W^{\left( v \right)}{}^\textsuperscript{\textit{T}} C^{\left( v \right)} W^{\left( v \right)} - 2 W^{\left( v \right)}{}^\textsuperscript{\textit{T}} D^{\left( v \right)}) 
\label{update_W1}
\end{equation}
\begin{equation}
\begin{cases}
C^{\left( v \right)} = \left(\alpha^{\left( v \right)}\right)^\gamma (\Theta^{\left( v \right)} X^{\left( v \right)} S^{\left( v \right)} H^{\left( v \right)} H^{\left( v \right)}{}^\textsuperscript{\textit{T}} S^{\left( v \right)}{}^\textsuperscript{\textit{T}} X^{\left( v \right)}{}^\textsuperscript{\textit{T}} \Theta^{\left( v \right)}{}^\textsuperscript{\textit{T}}) \\

D^{\left( v \right)} = \left(\alpha^{\left( v \right)}\right)^\gamma(\Theta^{\left( v \right)} X^{\left( v \right)} S^{\left( v \right)} H^{\left( v \right)} H^{\left( v \right)}{}^\textsuperscript{\textit{T}} S^{\left( v \right)}{}^\textsuperscript{\textit{T}} Y^\textsuperscript{\textit{T}} )
\end{cases}
\nonumber
\end{equation}

GPI\cite{GPI} can be employed to solve $W^{\left( v \right)}$ in Eq.~\eqref{update_W1}.

\subsection{Update $\Theta^{\left( v \right)}$  by fixing $W^{\left( v \right)}$, $\alpha^{\left( v \right)}$}
When $W^{\left( v \right)}$ and $\alpha^{\left( v \right)}$ are fixed and irrelevant terms are omitted, the following function for $\Theta^{\left( v \right)}$ can be written as:

\begin{equation} 
\begin{split}  
&\min_{\theta^{\left( v \right)}} \left(\alpha^{\left( v \right)}\right)^\gamma \left[ \lambda \theta^{\left( v \right)}{}^\textsuperscript{\textit{T}} R^{\left( v \right)} \theta^{\left( v \right)} + \right. \mathrm{Tr}(E^{\left( v \right)}F^{\left( v \right)}) \\
& \left. - 2 \mathrm{Tr}(\Theta^{\left( v \right)} X^{\left( v \right)} S^{\left( v \right)} H^{\left( v \right)} H^{\left( v \right)}{}^\textsuperscript{\textit{T}} S^{\left( v \right)}{}^\textsuperscript{\textit{T}} Y{}^\textsuperscript{\textit{T}} W^{\left( v \right)}{}^\textsuperscript{\textit{T}}) \right]  \\
& \hspace{2cm} \text{s.t. } \theta^{\left( v \right)}{}^\textsuperscript{\textit{T}} 1_{d^{\left( v \right)}} = 1, \quad \theta^{\left( v \right)} \geq 0
\end{split} 
\label{update_theta1}
\end{equation} 
where
$
\begin{cases}
E^{\left( v \right)} = H^{\left( v \right)}{}^\textsuperscript{\textit{T}} S^{\left( v \right)}{}^\textsuperscript{\textit{T}} X^{\left( v \right)}{}^\textsuperscript{\textit{T}} \Theta^{\left( v \right)}{}^\textsuperscript{\textit{T}} W^{\left( v \right)}{}^\textsuperscript{\textit{T}} \\

F^{\left( v \right)} = W^{\left( v \right)} \Theta^{\left( v \right)} X^{\left( v \right)} S^{\left( v \right)} H^{\left( v \right)}
\end{cases}
\nonumber
$

For diagonal matrix \( M \), \(\text{Tr}(MAMB) = M^T (A^T \circ B) M\) \cite{FSOR2}. The subproblem indicator matrix $S$ in Eq. (\ref{update_theta1}) can be reformulated as:

\[
\min_{\theta^{\left( v \right)}} \theta^{\left( v \right)}{}^\textsuperscript{\textit{T}} Q \theta^{\left( v \right)} - \theta^{\left( v \right)}{}^\textsuperscript{\textit{T}} g
\]
 \begin{equation}  
\text{s.t. } \theta^{\left( v \right)}{}^\textsuperscript{\textit{T}} 1_{d^{\left( v \right)}} = 1, \quad \theta^{\left( v \right)} \geq 0
\label{update_theta2}
\end{equation}  
Here,
\begin{equation}
\begin{cases}  
g = diag\left[2 \left(\alpha^{\left( v \right)}\right)^\gamma X^{\left( v \right)} S^{\left( v \right)} H^{\left( v \right)} H^{\left( v \right)}{}^\textsuperscript{\textit{T}} S^{\left( v \right)}{}^\textsuperscript{\textit{T}} Y{}^\textsuperscript{\textit{T}} W^{\left( v \right)}{}^\textsuperscript{\textit{T}}\right] \\
Q =  \left(\alpha^{\left( v \right)}\right)^\gamma\left[ \lambda R^{\left( v \right)}  +  O \circ (W^{\left( v \right)} W^{\left( v \right)}{}^\textsuperscript{\textit{T}})\right]
\end{cases}  
\nonumber
\end{equation}
where $O = X^{\left( v \right)} S^{\left( v \right)} H^{\left( v \right)} H^{\left( v \right)}{}^\textsuperscript{\textit{T}} S^{\left( v \right)}{}^\textsuperscript{\textit{T}} X^{\left( v \right)}{}^\textsuperscript{\textit{T}}$.

Eq. (\ref{update_theta2}) can be solved using ALM method \cite{ALM}.

\subsection{Update $\alpha^{\left( v \right)}$  by fixing $W^{\left( v \right)}$, $\Theta^{\left( v \right)}$}
When $W^{\left( v \right)}$ and $\Theta^{\left( v \right)}$ are fixed and irrelevant terms are omitted, the following function for $\alpha^{\left( v \right)}$ can be written as:

\[
\min_{\alpha^{\left( v \right)}} \sum_{v=1}^{\text{ch}}\left(\alpha^{\left( v \right)}\right)^\gamma U^{\left( v \right)}
\]
\begin{equation}  
\text{s.t. } \quad \sum_{v=1}^{\text{ch}} \alpha^{\left( v \right)} = 1, \quad 0 \leq \alpha^{\left( v \right)} \leq 1 
\end{equation} 
where $ U^{\left( v \right)} = \left[{ \left\| W^{\left( v \right)}{}^\textsuperscript{\textit{T}} \Theta^{\left( v \right)} X^{\left( v \right)} S^{\left( v \right)} H^{\left( v \right)} - Y S^{\left( v \right)} H^{\left( v \right)} \right\|_F^2}  \right. $\\
$ \left. +\lambda \theta^{\left( v \right)}{}^\textsuperscript{\textit{T}} R^{\left( v \right)} \theta^{\left( v \right)} \right] $.

This is a quadratic programming problem with a closed-form solution, which can be derived using the Karush-Kuhn-Tucker conditions and is given by: 
\begin{equation}
\alpha^{\left( v \right)} = \frac{\left(U^{\left( v \right)}\right){}^{\frac{1}{1-r}}}{ \sum_{v=1}^{\text{ch}} \left(U^{\left( v \right)}\right){}^\frac{1}{1-r}}
\label{alpha1}
\end{equation}

Algorithm~\ref{IDFS-MEC} outlines the detailed optimization steps for the objective Eq.~\eqref{Pf1}.

\begin{algorithm}[h]
\caption{Incomplete Depression Feature Selection with Missing EEG Channels (IDFS-MEC)}
\label{IDFS-MEC}
\begin{algorithmic}[1]
\Require EEG feature set $X = \left \lbrace X^{\left( 1 \right)}, X^{\left( 2 \right)}, ... , X^{\left( v \right)}, ... , X^{\left( ch \right)}\right \rbrace$ and its elements $X^{\left( v \right)} = [x_1, x_2, ... , x_d]^T$, depression label matrix $Y \in \mathbb{R}^{c \times n}$, missing-channel indicator $S = \left \lbrace S^{\left( 1 \right)}, S^{\left( 2 \right)}, ... , S^{\left( v \right)}, ... , S^{\left( ch \right)} \right \rbrace$ where $S^{\left( v \right)} \in \mathbb{R}^{n \times n}$, regularization parameters $\gamma$ and $\lambda$.
\State Initialize $\Theta^{\left( v \right)} \in \mathbb{R}^{d^{\left( v \right)} \times d^{\left( v \right)}}$ satisfying $\theta^{\left( v \right)}{}^\textsuperscript{\textit{T}} 1_{d^{\left( v \right)}} = 1$, where $\theta^{\left( v \right)} \geq 0$.
\State $ H = I_n - \frac{S^{\left( v \right)} \cdot 1_n \cdot 1_n^T}{\text{Tr}(S^{\left( v \right)} \cdot S^{\left( v \right)}{}^\textsuperscript{\textit{T}})}. $
\Repeat
\State Update $W^{\left( v \right)}$ via GPI.
\State Update $\Theta^{\left( v \right)}$ via ALM.
\State Update $\alpha^{\left( v \right)}$ via the Eq.(\ref{alpha1}) .
\Until convergence
\State \Return the regression matrix $W^{\left( v \right)} \in \mathbb{R}^{d^{\left( v \right)} \times c}$, the feature weight vector $\theta^{\left( v \right)}\in \mathbb{R}^{d^{\left( v \right)}}$, and the channel weighting vector $\alpha^{\left( v \right)}$.
\State Rank the EEG features via $ \bm{\theta}^{(v)}$ and $\alpha^{\left( v \right)}$.
\end{algorithmic}
\end{algorithm}

\begin{figure*}[!t]
    \centering
    \subcaptionbox{3-channel dataset (MODMA)\label{3accuracy}}{%
        \includegraphics[width=0.3\textwidth]{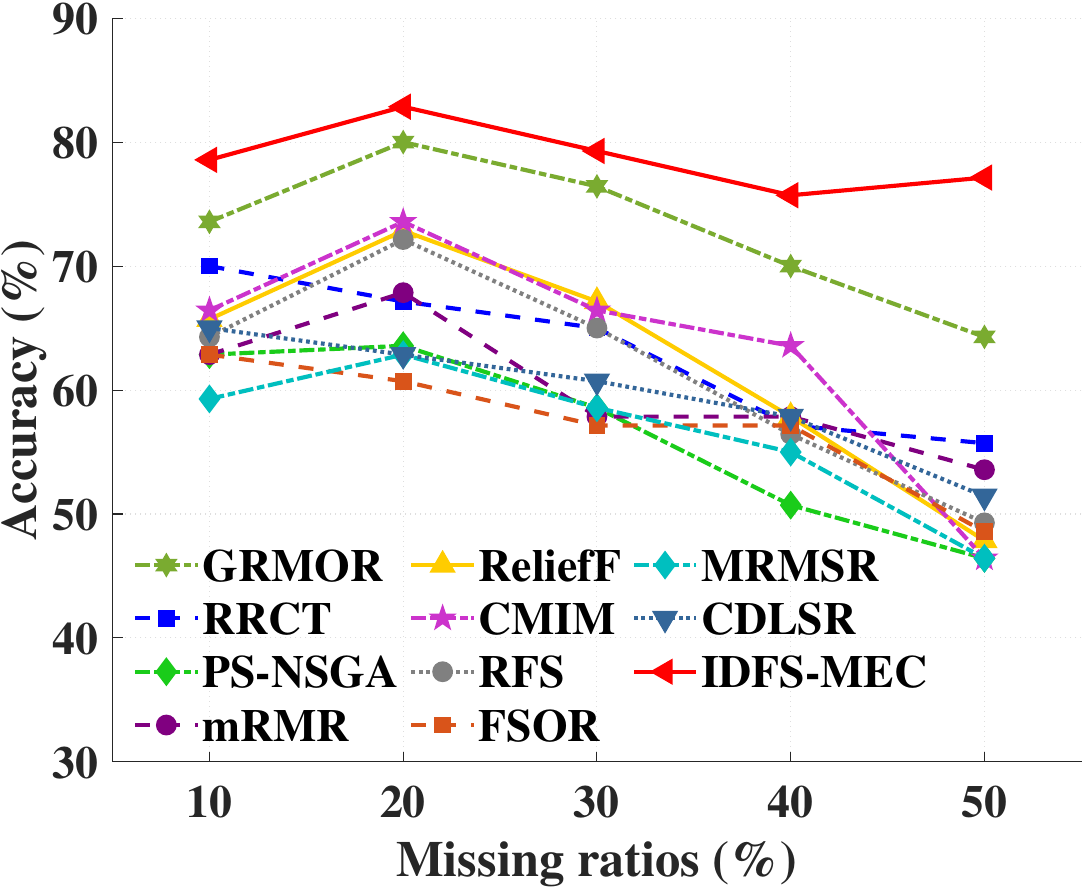}}
    \hfill
    \subcaptionbox{64-channel dataset (PRED-d003)\label{64accuracy}}{%
        \includegraphics[width=0.3\textwidth]{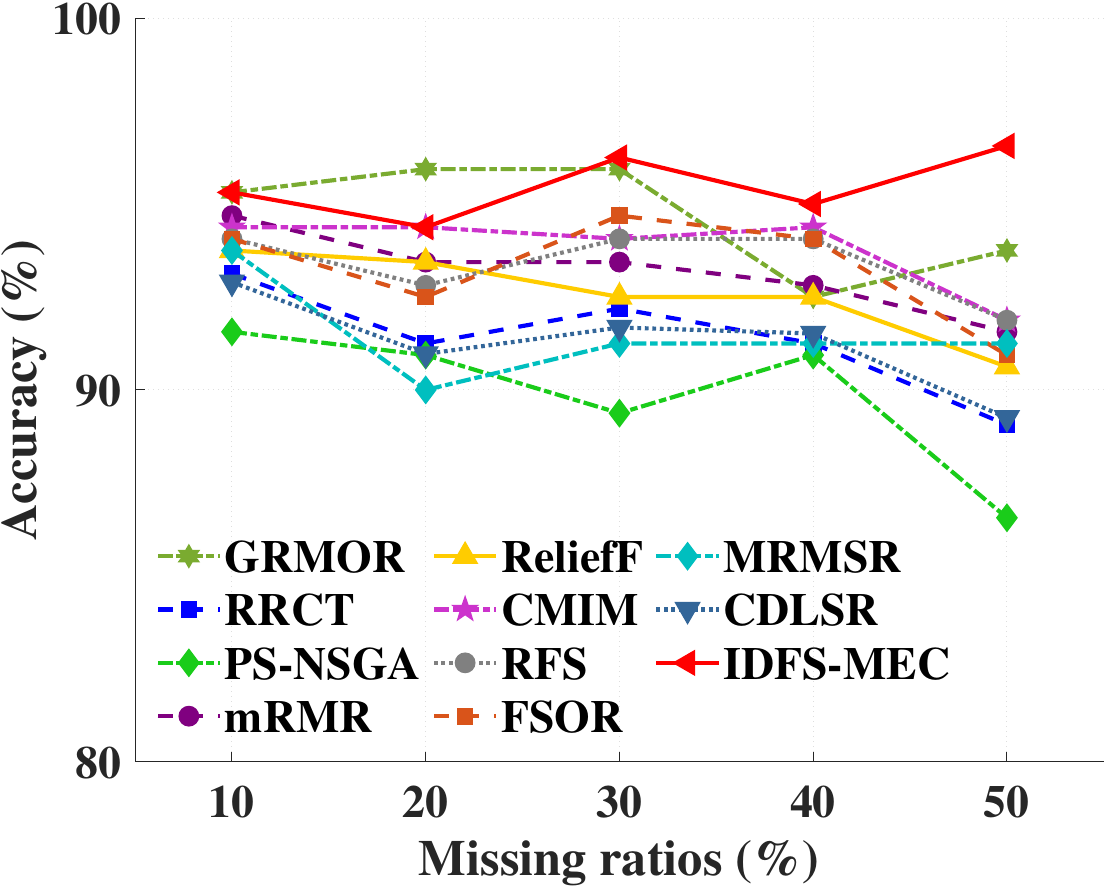}}
    \hfill
    \subcaptionbox{128-channel dataset (MODMA)\label{128accuracy}}{%
        \includegraphics[width=0.3\textwidth]{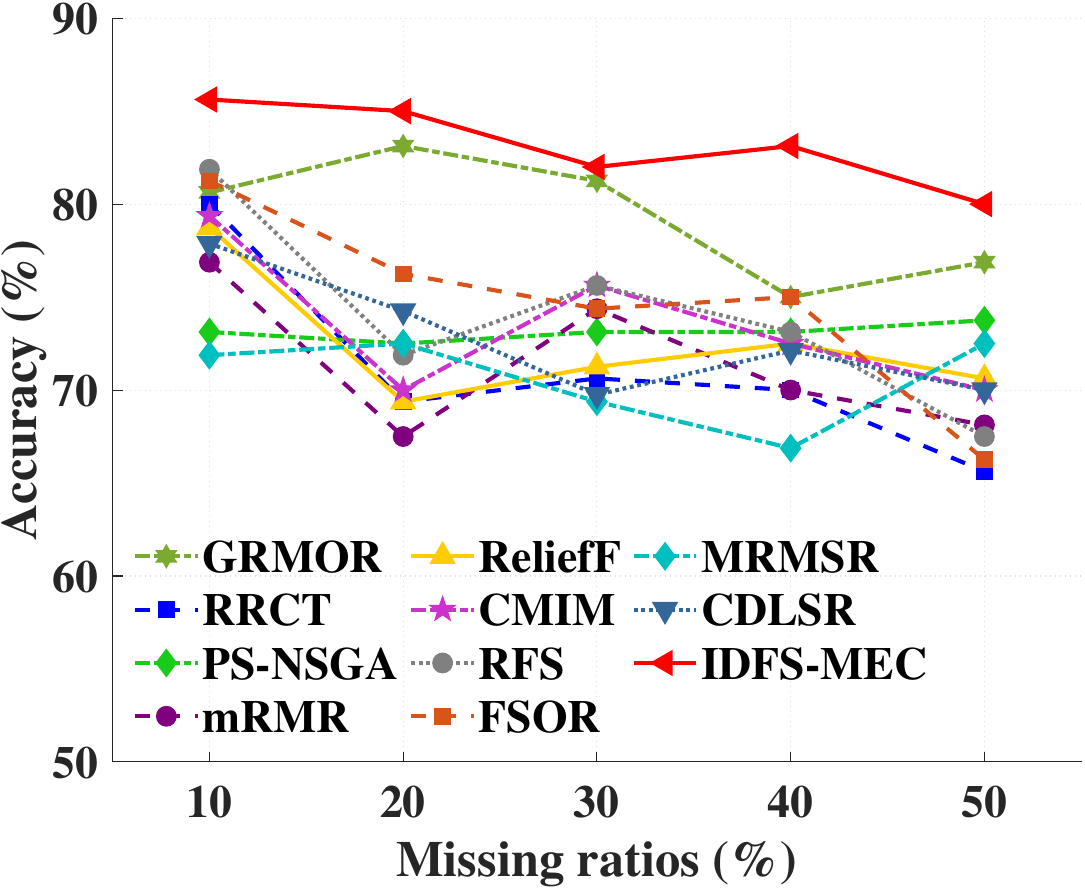}}
   % \captionsetup{font={small}}
    \caption{Classification accuracies on the 3-, 64-, and 128-channel EEG datasets.}
    \label{accuracygraph}
\end{figure*}

\begin{figure*}[!t]
    \centering
    \subcaptionbox{3-channel dataset (MODMA)\label{3parameter}}{%
        \includegraphics[width=0.29\textwidth]{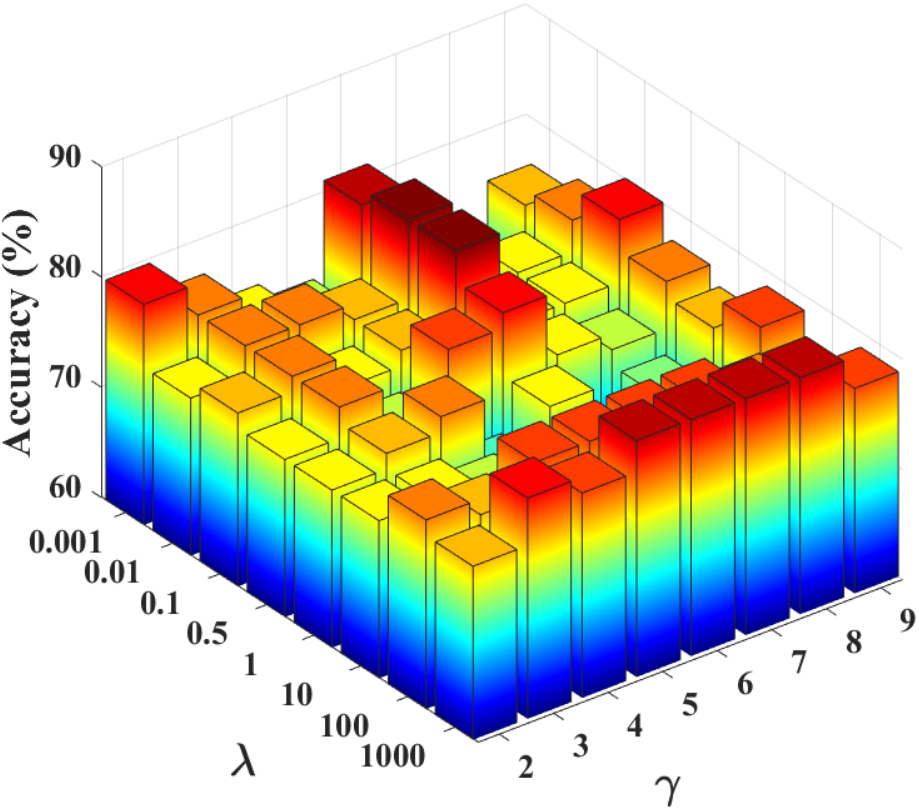}}
    \hfill
    \subcaptionbox{64-channel dataset (PRED-d003)\label{64parameter}}{%
        \includegraphics[width=0.29\textwidth]{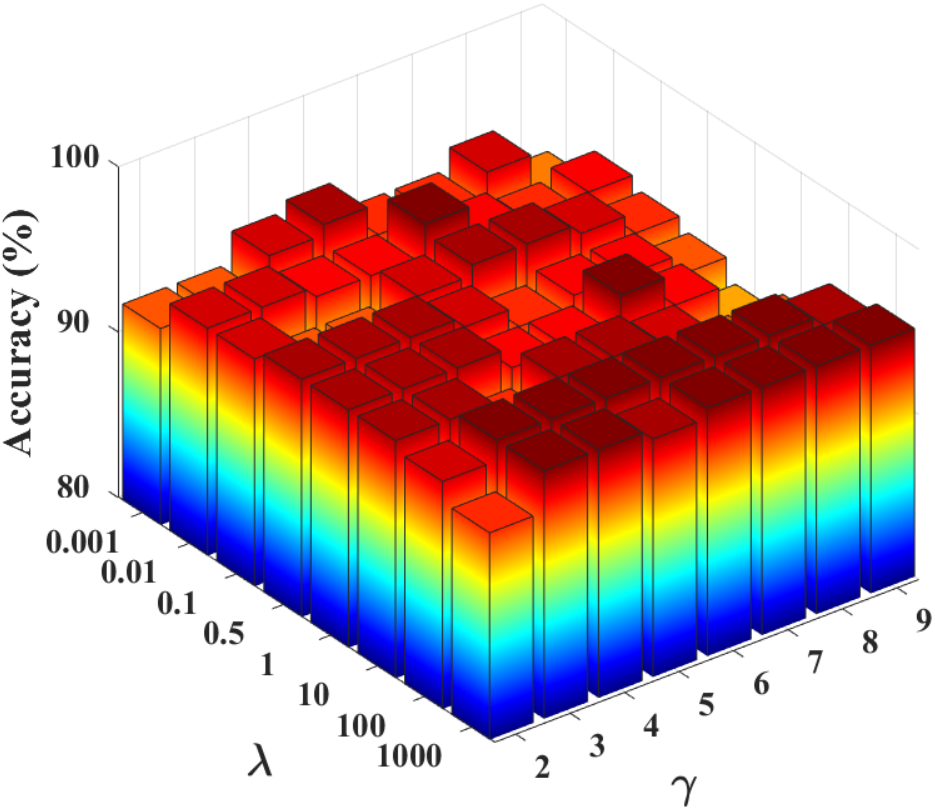}}
    \hfill
    \subcaptionbox{128-channel dataset (MODMA)\label{128parameter}}{%
        \includegraphics[width=0.29\textwidth]{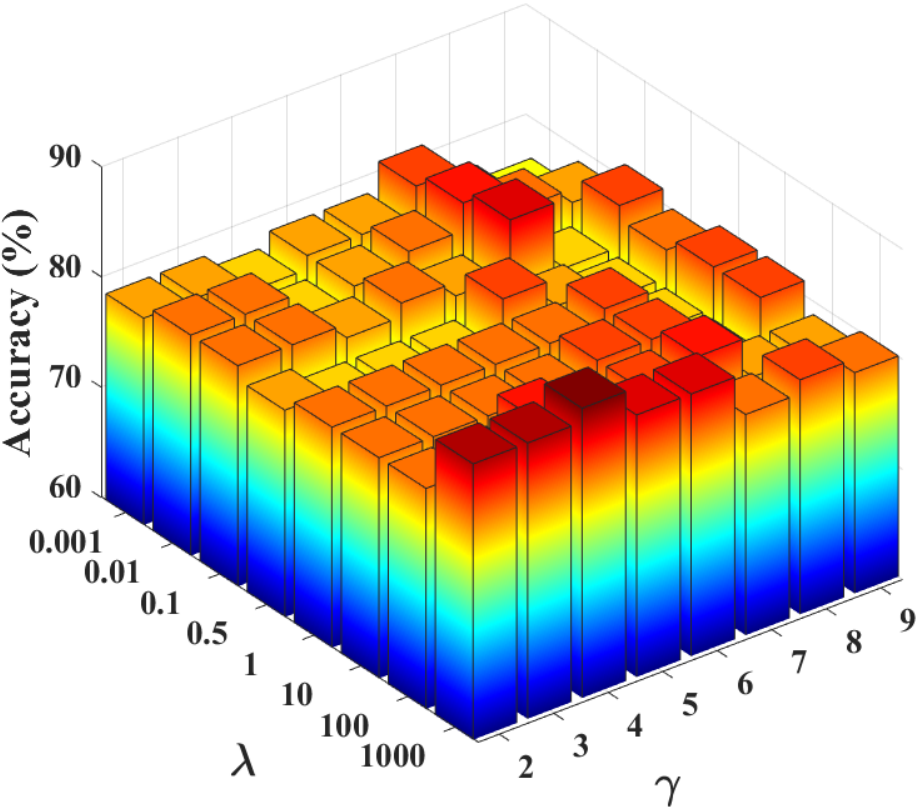}}
   % \captionsetup{font={small}}
    \caption{Average classification accuracies for different parameters on the 3-, 64-, and 128-channel EEG datasets.}
    \label{parameter_matrix}
\end{figure*}

\section{EXPERIMENTAL RESULTS AND DISCUSSION}

\subsection{Experimental Details}
MODMA (3- and 128-channel resting EEG, 250 Hz) \cite{EDD_3EEG,modma} and PRED-d003 (64-channel resting EEG, downsampled from 500 to 250 Hz) \cite{PRED} were adopted to analyze the performance of IDFS-MEC. The data were preprocessed with a 1-45Hz band-pass filter. A total of 15 EEG feature types were extracted in time and frequency domains \cite{XuGRMOR}. Time-domain features (mean, variance, non-stationary index, higher-order crossings, C0 complexity, spectral entropy, and Shannon entropy) captured time-series properties. To extract frequency-domain features, EEG signals were divided into five bands ($\delta$:1–4 Hz, $\theta$:4–8 Hz, $\alpha$:8–13 Hz, $\beta$:13–30 Hz, $\gamma$:30–45 Hz). Based on the above bands, derive features like differential entropy, absolute power ($AP$), and the ratio $AP_{\beta} / AP_{\theta}$ of $\beta$ and $\theta$ bands were extracted.

Ten advanced feature selection methods were compared with the proposed IDFS framework: Filter-based methods include ReliefF \cite{FSF1}, mRMR \cite{FSF6}, CMIM \cite{FSF7cmim}, MRMSR \cite{RW7}, and RRCT \cite{RRCT}. The wrapper-based technique PS-NSGA \cite{RW8} was employed. Embedded methods include RFS \cite{RFS}, FSOR \cite{FSOR}, GRMOR \cite{XuGRMOR}, and CD-LSR \cite{RW9}. Selected EEG features were fed into linear-core support vector machines (SVM) since it has been widely used in diagnosing depression \cite{svmgood}. All experiments were evaluated using ten-fold cross-validation. For IDFS-MEC, $\lambda$ is adjusted within the set $\left\{ 0.001, 0.01, 0.1, 1, 10, 100, 1000 \right\}$, while $\gamma$ is varied within $\left\{ 2, 3, 4, 5, 6, 7, 8, 9 \right\}$.

\subsection{Performance Analysis and Discussion}
\subsubsection{Feature Selection Porformance}
% \textbf{Feature Selection Porformance: }
Fig. \ref{accuracygraph} illustrates the optimal performance achieved by IDFS-MEC (red triangle) across the 3-, 64-, and 128-channel datasets under missing ratios ranging from 10\% to 50\% consistently. The accuracies of IDFS-MEC exhibit remarkable stability as the missing ratio increases, particularly on the challenging 3-channel dataset. The finding highlights the capability of IDFS-MEC to handle missing data under the limited-channel condition which is a scenario frequently encountered in clinical practice. Furthermore, IDFS-MEC continues to deliver robust and stable performance when processing incomplete data in high-density datasets (64- and 128-channels) with abundant information.

\subsubsection{Ablation Experiment}
% \textbf{Ablation Experiment: }
To evaluate the contribution of each component within IDFS-MEC, ablation experiments were conducted by removing the global redundancy minimization term $\theta^{\left( v \right)}{}^\textsuperscript{\textit{T}} R^{\left( v \right)} \theta^{\left( v \right)}$, the channel weighting parameter $\alpha^{\left( v \right)}$, and the missing-channel indicator matrix $S^{\left( v \right)}$. Table \ref{ablation} shows the ablation experimental results. The indicator matrix presents more significance in the 3-channel dataset, demonstrating the adaptability of IDFS-MEC in handling missing data under limited-channel conditions. In addition, channel weighting substantially enhances accuracies, underscoring the importance of adaptive channel selection within the IDFS-MEC framework.

\begin{table}[h]
    \centering  
    \begin{tabular}{c c c c}    
        \hline
        \textbf{Method} & \textbf{3-ch} & \textbf{64-ch} & \textbf{128-ch} \\
        \hline  
        abl\_Redundancy\_Minimization & 69.71 & 90.56 & 77.50 \\
        abl\_Channel\_Weight & 71.85 & 88.31 & 70.50 \\
        abl\_Missing\_Channel\_Indicator & 74.57 & 91.48 & 84.50 \\
        \textbf{IDFS-MEC} & \textbf{78.72} & \textbf{95.50} & \textbf{87.50} \\
        \hline  
    \end{tabular}
   % \captionsetup{font={small}}
\caption{Ablation experiments on average accuracies across 5 missing rates for 3-, 64-, and 128-channel EEG datasets.}
\vspace{-9pt}
\label{ablation}  
\end{table}

\subsubsection{Parameter Sensitivity Analysis}
% \textbf{Parameters Sensitivity: }
Fig. \ref{parameter_matrix} illustrates the impact of different combinations of parameters on the feature selection performance of IDFS-MEC. In the 3-channel dataset, IDFS-MEC exhibits higher parameter sensitivity, with clear performance peaks around $\lambda=100,1000$ and $\gamma=6$, suggesting that the method can adaptively select parameters to adjust data characteristics. As the number of channels increases, this sensitivity diminishes, indicating that IDFS-MEC achieves greater stability under higher-density EEG settings.

\subsubsection{Computational Costs}
% \textbf{Computation Cost: }
Table \ref{Compu_cost} provides detailed results on the average computation times for representative methods. All evaluations were conducted in MATLAB on a Windows 11 X64 system, equipped with an AMD Ryzen 7 4800H processor and 16GB of RAM. As shown in Fig. \ref{accuracygraph} and Table \ref{Compu_cost}, IDFS-MEC achieves optimal performance in depression-related feature selection while requiring relatively low computational costs, demonstrating the effectiveness of the proposed method.

\begin{table}[h]
    \centering
    \begin{tabular}{c|cccc}
    \hline
    \textbf{Type} & \textbf{Method} & \textbf{3-ch} & \textbf{64-ch} & \textbf{128-ch}\\
    \hline
    \multirow{5}{*}{Filter} & mRMR & 0.008 & 0.370 & 1.484\\
    & CMIM & 0.003 & 0.093 & 0.364\\
    & ReliefF & 0.090 & 0.284 & 1.181\\
    & RRCT & 0.139 & 4.468 & 15.317\\
    & MRMSR & 1.288 & 39.401 & 256.088\\
    \hline
    Wrapper & PS-NSGA & 1.740 & 24.633 & 51.938\\
    \hline 
    \multirow{5}{*}{Embedded} & RFS & 0.017 & 0.055 &0.063\\
    & FSOR & 0.009 & 0.063 &0.066\\
    & GROMR &  0.275 & 8.235 &47.173\\
    & CDLSR & 0.014 & 6.063 & 4461.149\\
    & \textbf{IDFS-MEC} & \textbf{0.143} &\textbf{0.682} &\textbf{6.256}\\
    \hline
    \end{tabular}
  %  \captionsetup{font={small}}
    \caption{The comparison of average computational time (seconds).}
    \vspace{-10pt}
    \label{Compu_cost}
\end{table}

\section{Conclusions and future work}
In the deployment of brain-computer interfaces for practical depression applications, environmental noise or sensor drift often leads to partial loss of EEG channel data. To solve the problem, this study introduced a novel EEG feature selection framework to address redundancy and missing channel challenges in EEG-based depression detection. By combining weighted orthogonal regression, adaptive channel weighting learning, and global redundancy minimization learning, IDFS-MEC consistently outperforms among ten advanced feature selection methods across 3-, 64-, and 128-channel datasets, with strong robustness to incomplete EEG channels and relatively low computational costs. 

Future research should prioritize additional modalities and datasets with various degree of depression to rigorously assess IDFS-MEC across diverse and complex device environments.

\section{Acknowledgments}
This work was funded by: the National Natural Science Foundation of China under Grant No.62502019, the Beijing Natural Science Foundation under Grant No.4244087, China Postdoctoral Science Foundation under Grant No.2022M720332, Beijing Postdoctoral Research Foundation under Grant No.2023-zz-85, and Chaoyang Postdoctoral Research Foundation under Grant No.2023ZZ-004.

\newpage
\bibliographystyle{IEEEbib}
\bibliography{strings,refs}

\end{document}